\documentclass[sigchi]{acmart}

\usepackage{float}
\usepackage{tabulary,xcolor}
\usepackage{algorithm, algorithmic}
\usepackage{graphicx}
\usepackage{setspace}
\usepackage{adjustbox}
\usepackage{multirow}
\usepackage{subcaption}

\def\BibTeX{{\rm B\kern-.05em{\sc i\kern-.025em b}\kern-.08emT\kern-.1667em\lower.7ex\hbox{E}\kern-.125emX}}

\copyrightyear{2018}
\acmYear{2019}
\setcopyright{acmlicensed}
\acmConference[]{}{}{}
\acmBooktitle{}
\acmPrice{}
\acmDOI{}
\acmISBN{}

\begin{document}

\title{GPU accelerated matrix factorization of large scale data using block based approach}

\author{Prasad G Bhavana}
\email{prasadb@ieee.org}
\author{Vineet Padmanabhan}
\email{vineetcs@enet.uohyd.ac.in}
\affiliation{%
  \institution{University of Hyderabad}
  \streetaddress{School of Computer and Information Sciences}
  \city{Hyderabad}
  \country{India}
  \postcode{500046}
}

\renewcommand{\shortauthors}{Prasad Bhavana and Vineet Padmanabhan}

\begin{abstract}
Matrix Factorization (MF) on large scale data takes substantial time on a Central Processing Unit (CPU). While Graphical Processing Unit (GPU)s could expedite the computation of MF, the available memory on a GPU is finite. Leveraging GPUs require alternative techniques that allow not only parallelism but also address memory limitations. Synchronization between computation units, isolation of data related to a computational unit, sharing of data between computational units and identification of independent tasks among computational units are some of the challenges while leveraging GPUs for MF. We propose a block based approach to matrix factorization using Stochastic Gradient Descent (SGD) that is aimed at accelerating MF on GPUs. The primary motivation for the approach is to make it viable to factorize extremely large data sets on limited hardware without having to compromise on results. The approach addresses factorization of large scale data by identifying independent blocks, each of which  are factorized in parallel using multiple computational units. The approach can be extended to one or more GPUs and even to distributed systems. The RMSE results of the block based approach are with in acceptable delta in comparison to the results of CPU based variant and multi-threaded CPU variant of similar SGD kernel implementation. The advantage, of the block based variant, in-terms of speed are significant in comparison to other variants.
\end{abstract}


\maketitle

\section{Introduction}
Matrix factorization is a simple and effective machine learning technique used for discovering latent/hidden factors that can approximately explain large volumes of observed behavior. In the recommender systems domain, the observed behaviours are the ratings that are associated with two distinct dimensions i.e \emph{users} and \emph{items}. MF is not limited to recommender systems alone and has been leveraged in domains like text mining, astronomy, data compression, image processing etc. to solve varieties of problems. The fundamental approach is to take into account large volumes of observation so as to derive patterns that can be used for prediction/classification. With time, as more and more observations get accumulated, parallel processing techniques based on leveraging partial data elements are needed. Block based, GPU accelerated approach to matrix factorization (BGMF) provides an alternative method that is effective in terms of memory management and provides a way to parallelism in computation. \\

In~\cite{Tan2018MatrixFO} a GPU accelerated matrix factorization based on SGD as well as approximate ALS technique has been proposed. Optimization of memory within GPU is also taken care in~\cite{Tan2018MatrixFO}. GPU accelerated Non-Negative Matrix Factorization (NNMF) for CUDA (Compute Unified Device Architecture) capable hardware has been proposed in~\cite{koitka2016nmfgpu4r}. Similarly~\cite{10.1007/978-3-642-31178-9_15} proposed NNMF with GPU acceleration for text mining purposes. From the literature it can be found that various matrix factorization based approaches have been proposed which includes parallel as well distributed frameworks for scaling up the factorization process~\cite{NIPS20114486}, ~\cite{Zhang:2013:LMF:2488388.2488520}, ~\cite{Du2017}, ~\cite{7727641}, \cite{DBLP:journals/corr/SchelterSZ14}, \cite{DBLP:conf/kdd/GemullaNHS11}, \cite{Yun:2014:NNS:2732967.2732973}, \cite{DBLP:conf/edbt/LiTS13}, \cite{Zhuang:2013:FPS:2507157.2507164}, \cite{DBLP:conf/nips/RechtRWN11}, \cite{DBLP:conf/kdd/OhHYJ15}, \cite{DBLP:conf/icdm/YuHSD12}, \cite{DBLP:conf/pakdd/ChinZJL15}. Even then it is needless to say that all these techniques need further adaptations for GPU acceleration. BGMF, as proposed in this work, provides a framework that can be used by other matrix factorization optimization approaches for scaling up as well providing parallelism and distributed computation along with GPU acceleration. BGMF relies on kernel implementation that takes partial data elements to achieve MF and hence currently it is implemented using Stochastic Gradient Descent (SGD). However, the approach can be extended to approximate techniques based on Alternating Least Squares (ALS) approach that uses partial data.

Let $X \in \mathbb{R}^{\{n\times m\}}$ be a matrix with $n$ users and $m$ observations per user. Typically, in a regular MF approach, the data matrix $X$ can be approximated as $X \approx UV^T$; where user latent factor matrix $U \in \mathbb{R}^{\{n \times k\} }$, for some given $k$ dimensional latent feature vectors, is derived from the given observations. Similarly, the item latent factor matrix $V \in \mathbb{R}^{\{m\times k\} }$, for $k$ latent features, is derived from given observations. The optimization problem of MF can be represented as:

\begin{equation}
\min_{U, V} \sum_{x_{ij}} ((x_{ij} - {u_i^T}{v_j})^2 + \beta(||{u_i}||^2+ || v_j||^2) )
\label{eq:1}
\end{equation}

where $\beta$ is the regularization parameter. The update equation for $i^{th}$ row of $U$ for SGD convergence is:

\begin{equation}
    u_{ik}^{'}
    = u_{ik} + \alpha\big(2e_{ij}v_{kj} - \beta u_{ik} \big)
\end{equation}
where $\alpha$ is the learning parameter. Similarly the update equation for \ensuremath{j^{th}} column of \ensuremath{V} for SGD convergence is given as:

\begin{equation}
    v_{kj}^{'} = v_{kj} + \alpha\big(2e_{ij}u_{ik} - \beta v_{jk} \big)
\end{equation}

The block matrix representation~\eqref{eq:datax} of the data matrix is based on concatenation of blocks/sub-matrices of equal dimension as shown below:

\begin{equation}
X = \begin{bmatrix}
	\underline{X}_{11} & \underline{X}_{12} & \dots & \underline{X}_{1j} & \dots  & \underline{X}_{1J} \\
	\vdots & \vdots & \dots & \vdots & \ddots & \vdots \\
	\underline{X}_{i1} & \underline{X}_{i2} & \dots & \underline{X}_{ij} & \dots  & \underline{X}_{iJ} \\
	\vdots & \vdots & \dots & \vdots & \ddots & \vdots \\
	\underline{X}_{I1} & \underline{X}_{I2} & \dots & \underline{X}_{Ij} & \dots  & \underline{X}_{IJ}
\end{bmatrix}\label{eq:datax}
\end{equation}
\\

Since the data is assumed to be i.i.d for each user and respective item, Matrix Factorization does not constrain the sequence of processing of ratings to derive latent features. The technique results in optimized results when each of the data elements are processed once per iteration. Block based matrix factorization approach decomposes any given matrix into a block matrix and each sub-matrix is independently factorized, while the resultant latent feature sub-matrices are reused for factorization of other relevant sub-matrices. The overall process is repeated till convergence. The GPU acceleration is achieved by dedicating different independent blocks to different GPU cores and each block is factorized in parallel. Approximate, but comparable to global optima, results can also be achieved with factorization of each block for more than one iteration as detailed in Section \ref{sssec:block}.

The objective of block based MF is to find $\underline{U_i}, \underline{V_j}$ latent blocks for $\underline{X_{ij}}$ data block such that
$\underline{X_{ij}} \approx \underline{U_i}.{\underline{V_j}}^T$. Let $\underline{E_{ij}}$ represent the deviation of estimate $\underline{X_{ij}}^\prime$ from actual ($\underline{X_{ij}}$). Hence sum of squared deviations ($\mathcal{E}$), for a block, can be represented as:

\begin{equation}
    \displaystyle{\mathcal{E}} = ||\underline{X_{ij}} -\underline{X_{ij}}^\prime||^2 = ||\underline{X_{ij}} - \underline{U_{i}} {\underline{V_j}}^T||^2
\label{eq:8}
\end{equation}

Our objective now is to find $\underline{U_i}, \underline{V_j}$ such that the sum of squared deviations is minimal. The optimum value for $g^{th}$ latent feature of $\underline{U_i}$ block can be arrived by minimizing Equation (\ref{eq:8}) w.r.t $\underline{U_i}$ as:

\begin{equation}
\displaystyle{\frac{\partial}{\partial \underline{U_i}}_{*g}\mathcal{E}} = -2 ( \underline{X_{ij}} - \underline{X_{ij}}^\prime ).\underline{V_j}_{*g}
\label{eq:9}
\end{equation}

Similarly, optimal value for for $g^{th}$ latent feature of $\underline{V_j}$ can be arrived by minimizing Equation (\ref{eq:8}) w.r.t $\underline{V_j}$ as:

\begin{equation}
\displaystyle{\frac{\partial}{\partial \underline{V_j}}_{*g}\mathcal{E}} = -2{( \underline{X_{ij}} - \underline{X_{ij}}^\prime)^T}\underline{U_i}_{*g}
\label{eq:10}
\end{equation}

Using the block level gradients for a latent feature, the update equation for $g^{th}$ column of $\underline{U_i}$ can be arrived as:

\begin{equation}
\displaystyle{\underline{U_i}_{*g}^\prime} = \underline{U_i}_{*g} + \alpha \frac{\partial}{\partial \underline{U_i}_{*g}} \mathcal{E} = \underline{U_i}_{*g} + 2\alpha\underline{E}_{ij}\underline{V_j}_{*g}
\label{eq:11}
\end{equation}

Similarly, the update equation for $g^{th}$ column of $\underline{V_j}$:

\begin{equation}
\displaystyle{\underline{V_j}_{*g}^\prime} = \underline{V_j}_{*g} + \alpha \frac{\partial}{\partial \underline{V_j}_{*g}} \mathcal{E} = \underline{V_j}_{*g} + 2\alpha\underline{E_{ij}}^{T}\underline{U_i}_{*g}
\label{eq:12}
\end{equation}

Factoring in regularization term into the optimization Equation (\ref{eq:8}) gives us:

\begin{equation}
\mathcal{E} = ||\underline{X_{ij}} - \underline{X_{ij}}^\prime||^2 + \frac{\beta}{2} (\sum ||\underline{U_i}||^2 + ||\underline{V_j}||^2 )
\label{eq:13}
\end{equation}

Now, the update Equation (\ref{eq:11}) changes accordingly with minimization of Equation (\ref{eq:13}) w.r.t to $\underline{U_i}$ to include regularization term:

\begin{equation}
\displaystyle{\underline{U_i}_{*g}^\prime} = \underline{U_i}_{*g} + \alpha(2\underline{E_{ij}}\underline{V_j}_{*g} - \beta \underline{U_i}_{*g})
\label{eq:14}
\end{equation}

Similarly, the Equation (\ref{eq:12}) for update of $g^{th}$ column of $\underline{V_j}$ changes accordingly to include regularization term:

\begin{equation}
\displaystyle{\underline{V_i}_{*g}^\prime} = \underline{V_j}_{*g} + \alpha(2\underline{E_{ij}}^{T}\underline{U_i}_{*g} - \beta \underline{V_j}_{*g})
\label{eq:15}
\end{equation}

\subsection{Parallelism}

\begin{figure}[!h]
  \includegraphics[width=\linewidth]{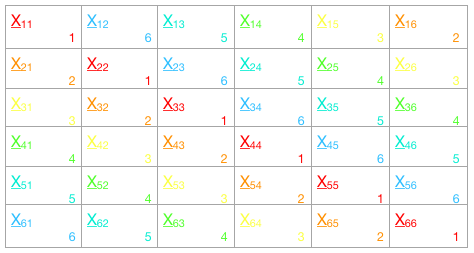}
  \caption{Example of parallel BMF}
  \label{fig:bmf}
\end{figure}

$U_i, V_j$ matrices can be derived simultaneously for multiple blocks by identifying the data blocks whose latent factors do not depend on each other and dedicating a GPU core (Computation unit) for each such block. Figure~\ref{fig:bmf} shows one such example with a $6\times6$ block matrix wherein blocks are represented with the same color (and a number placed at lower right corner of the cell) can be simultaneously factorized; i.e blocks on the diagonal are processed in first parallel step and then the factorization is moved on to the blocks below them while considering the entire column as a loop starting from the diagonal block.

\subsection{Algorithm}

\begin{spacing}{0.8}
\begin{algorithm}[!h]
\begin{algorithmic}

\REQUIRE{Input: Data matrix $X \in \mathbb{R}^{\{n \times m\} }$, number of features $k$}
\item[]

\STATE{Initialize: latent feature matrices with random values $U \in \mathbb{R}^{\{n \times k\}}, V \in \mathbb{R}^{\{m \times k\} }$}

\STATE{Let $I, J$ be two constants such that $X$ is represented by $I \times J$ number of sub-matrices}
\item[]

\STATE{Represent data matrix $X$ as block matrix with sub-matrices $X_{i j}$ where $i \in 1..I$ and $j \in 1..J$}
\item[]
\STATE{Similarly, represent feature matrices $U, V$ as block matrix with sub-matrices $U_i, V_j$ where $i \in 1..I$ and $j \in 1..J$}
\item[]

\STATE{Let CPU\_STEPS, GPU\_STEPS be constants representing maximum iterations of MF in CPU and GPU. Let $\alpha$ be the learning rate, $\beta$ be the regularization factor and $\delta$ the minimum deviation of error between iterations}
\item[]

\FOR{step 1 to CPU\_STEPS}{

	\FOR{ each column $j$ in block matrix $X$ }{

		\item[]
		\STATE{$l \gets  mod( (i+step), I)$}
		\STATE{$m \gets  mod( (j+step), J)$}
		\item[]
		\STATE{Wait if there is a thread running on same row as $l$}
		\item[]
		\STATE{Invoke thread: $U_l, V_m \gets gpumf(X_{l m}, U_l, V_m$, $k$, $\alpha$, $\beta$, GPU\_STEPS)}
		\\
		\item[]
		\item[]

		\COMMENT{$gpumf()$ function is SGD based MF algorithm implementation, at block level, for GPU\_STEPS and derives k latent features. The function is executed on one GPU core per call, while the function call itself is executed in a multi-threaded mode}
		\item[]
	}
	\ENDFOR
	\STATE{Terminate convergence if RMSE improvement is $< \delta$ for entire matrix}

}
\ENDFOR
\item[]
\RETURN{ Latent feature matrices $U \in \mathbb{R}^{\{n \times k\} }, V \in \mathbb{R} ^{\{m \times k\} }$}
\item[]
\item[]
\caption{Block based GPU accelerated MF}
\label{alg:one}
\end{algorithmic}
\end{algorithm}
\end{spacing}

A broad outline of the approach is given in Algorithm~\ref{alg:one}. The $gpumf()$ function in the algorithm encapsulates GPU implementation of SGD based MF. The function is implemented to factorize the given block for GPU\_STEPS number of times. The two for loops, iterating CPU\_STEPS, GPU\_STEPS (number of times for each sub-matrix), together contribute to $I$ (number of iterations) which is independent of data elements and hence can be considered as constant. Hence the time complexity of the algorithm remains same as MF. Considering the block size as constant, the space required for factorizing each block is $\approx ( n\times m + n\times k + k\times m) \times c$ for some constant $c$. The space required for factorization of block can be explained as the space taken up the data block and by the two latent factor blocks. As the blocks are of fixed dimension, the space complexity at a block level is constant. As the total computational units on a GPU are constant, and as the number of blocks factorized in parallel are constant, hence the space complexity of the algorithm remains $\approx O(c)$ (constant)
\\
The implementation of the BGMF algorithm is made available for public on Github at~\footnote{https://github.com/17mcpc14/blockgmf} along with code for the experiments presented in section~\ref{sssec:analysis}.

\subsection{CPU vs GPU iterations}\label{sssec:block}

Frequent transfer of blocks to and from GPU is a major cost in GPU computation. Factorizing each block for more than one iteration reduces the number of data transfers. Factorization is optimal when each block is factorized once and the respective $\underline{U}_i$, $\underline{V}_j$ latent features are passed on to factorize the following sub-matrices. It is also the case that approximate but nearly equivalent to optimal results are observed when each block is factorized more than once per step.

\begin{figure}[!t]
	\includegraphics[width=\linewidth]{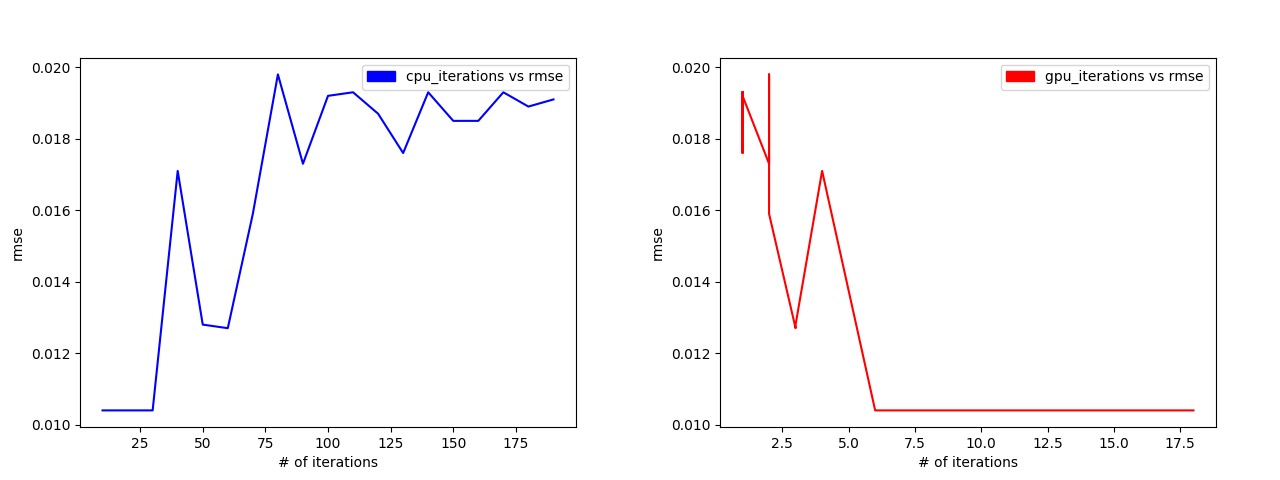}
	\caption{Comparison of different combinations of GPU, CPU iterations}
	\label{fig:cpugpuerrorgraph}
\end{figure}

Figure~\ref{fig:cpugpuerrorgraph} demonstrates differences in convergence at different combinations of CPU and GPU iterations while keeping the overall iterations (CPU\_STEPS $\times$ GPU\_STEPS) constant at 200 iterations. With the increase in GPU iterations, the CPU iterations come down and vice-versa; i.e the CPU\_STEPS variable is incremented from 1 to 100 and simultaneously GPU\_STEPS variable is decreased from 200 to 1 so that the total iterations (CPU\_STEPS $\times$ GPU\_STEPS) is kept constant at 200 for this experimentation setup. The data matrix used for the experiment is a dense matrix with size $1000\times1000$ with $k$ value taken as 30.

 As the GPU iterations improve the convergence at an individual block level, the RMSE also improves at an individual block level. The CPU iterations improve the transfer of latent features between the blocks thereby resulting in improvements in RMSE convergence across the entire matrix. In the above case 25 GPU iterations and 8 CPU iterations are optimal combination.

Hence, under constrained (overall number of iterations) scenario, identifying the right combination of CPU, GPU iterations is the key to achieving quicker convergence, without compromising outcome. While the combination of CPU, GPU iterations change with the overall number of iterations, in our observation the combination also differs with the nature of data. When the block sizes are relatively large and when the blocks are true representative sample of the overall data, the interdependence between the blocks is reduced. The latent features for each block are nearly equal to the latent features corresponding to other blocks and thus the results of factorization of just few blocks is nearly equivalent to the factorization of the entire matrix.

 Clearly there exists an alternative approach that gives better performance than that of the regular approach of considering each element, block once per iteration for factorization. The experiments detailed in Section \ref{sssec:block} (Block level factorization) dwells further in this direction.

\section{Experiments}\label{sssec:analysis}

The experiments mentioned below are conducted on a shared hardware with Intel(R) Xeon(R) CPU E5-2640 v3 @ 2.60GHz and with Nvidia Tesla M40 GPU with 24GB of CPU memory and 8GB of GPU memory. The programming environment is python that leverages NVIDIA CUDA with driver version 8.0 and with PyCUDA application programming interface version 1.8.

\subsection{Performance Analysis}\label{sssec:time}

The experiments in this section are conducted on a pseudo data set (generated with random numbers with values ranging between 1 to 30 ) of $1024\times1024$ matrix. The $\alpha=0.0001$, $\beta=0.01$ are the learning, regularization parameters while $\delta=0.01$ as the minimum difference of error between steps considered for early termination. The $k$ (number of latent features approximated) taken for the experiments is 10.

The experiments are conducted on 4 different matrix factorization variants implemented based on stochastic gradient descent approach:
\begin{itemize}
    \item CMF - CPU based MF
    \item CPMF - Multi-threaded CPU based MF with 1024 threads synchronized
    \item GMF - GPU based MF with 1024 cores synchronized
    \item BGMF - Block based GPU MF with 1024 cores/blocks processed in parallel
\end{itemize}

\begin{table}[!h]
		\begin{tabular}{|l|c|c|c|}
			\hline

Variant  & Seconds/Iteration & Iterations & RMSE \\ \hline
CMF	& 37.133 & 5 & 8.568\\ \hline
CPMF & 26.589 & 6 & 8.569 \\ \hline
GMF & 3.680 & 6 & 8.569\\ \hline
BGMF & 3.067 & 6 & 8.569\\ \hline
	\end{tabular}
\caption[t2]{ Performance comparison of Block based GPU MF against CMF, CPMF, GPMF}
\label{tab:time}
\end{table}

\begin{figure}[!t]
	\includegraphics[width=\linewidth]{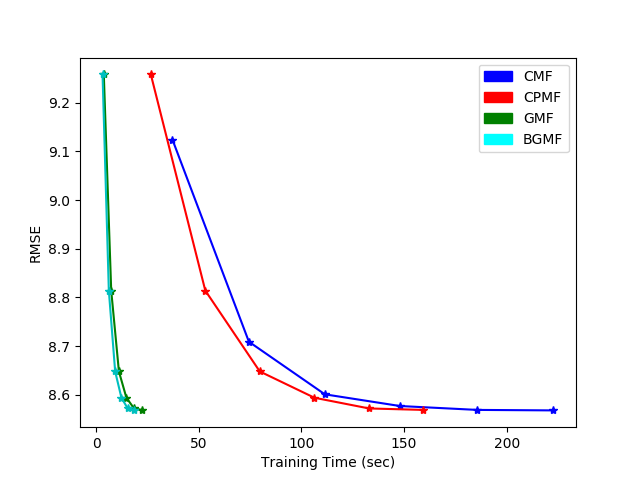}
	\caption{Comparison of CPU MF (CMF), Multi-threaded CPU MF (CPMF), Atomic GPU MF (GPMF), Block based GPU MF (BGMF) on a pseudo random data set. }
	\label{fig:timegraph}
\end{figure}

A comparison of the results related to the 4 MF variants as mentioned above are detailed in Table~\ref{tab:time}. Figure~\ref{fig:timegraph} demonstrates the comparative performance advantage of GPU variants of MF over CPU variants. The computational cost on CPU is substantially more than that on any of the GPU variants, including the multi-threaded parallel CPU MF. With the GPU variant of MF wherein  synchronization between cores is considered, as we increase the number of cores, the synchronization becomes a challenge and thereby incurs additional wait time. This is addressed by block based variant which drastically reduces the synchronization task to launching one thread (processing single block) per column within the block matrix.

\subsection{Comparison against standard data sets}

\begin{figure*}[h]
	\includegraphics[width=\textwidth]{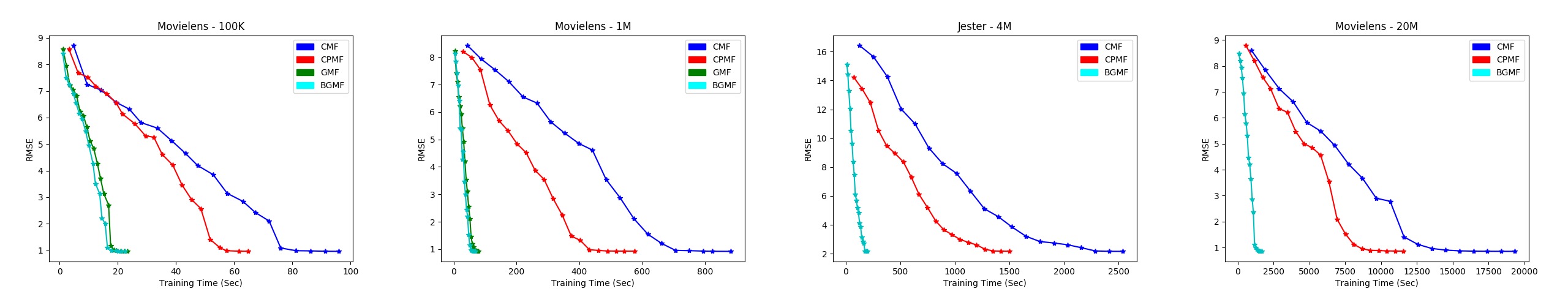}
\caption{Comparison of CPU, Block based GPU MF (BGMF) against standard data sets}
\label{fig:standard}
\end{figure*}

Table~\ref{tab:standard} and Figure \ref{fig:standard} lists the comparative analysis of time taken per iteration, number of iterations to converge and Test RMSE for the four different variants of MF listed in Section \label{sssec:time}.  The data sets used are MovieLense 100K~\footnote{http://grouplens.org/datasets/movielens/100k/}, MovieLense 1M~\footnote{http://grouplens.org/datasets/movielens/1m/}, MovieLense 20M~\footnote{https://grouplens.org/datasets/movielens/20m/} and Jester~\footnote{http://www.ieor.berkeley.edu/~goldberg/jester-data/}. All the experiments in Table~\ref{tab:standard} are conducted on standard hardware as mentioned at the beginning of the section~\ref{sssec:analysis}. The $k$ (dimension of latent feature vectors) value taken for all the experiments is 30 and $\alpha$ = 0.0001, $\beta$ = 0.01 are the learning, regularization parameters. The RMSE values of the experiments may not be comparable to that of the results from other papers, as the focus of the experiments are primarily to compare performance and convergence of CPU, GPU and block variants of MF.\\

\begin{table}[!h]

\resizebox{\linewidth}{!}
{

\begin{tabular}{c|c|r|r|r|r|}
\cline{2-6}
\multicolumn{1}{l|}{}                                                                                           & \multicolumn{1}{l|}{Data set --\textgreater{}} & \multicolumn{1}{c|}{\begin{tabular}[c]{@{}c@{}}MovieLense\\  100K\end{tabular}} & \multicolumn{1}{c|}{\begin{tabular}[c]{@{}c@{}}MovieLense \\ 1M\end{tabular}} & \multicolumn{1}{c|}{\begin{tabular}[c]{@{}c@{}}Jester \\ 4M\end{tabular}} & \multicolumn{1}{c|}{\begin{tabular}[c]{@{}c@{}}MovieLense \\ 20M\end{tabular}}\\ \cline{2-6}
\multicolumn{1}{r|}{} & n (users) & 1000 & 6000 & 73421 & 138000
\\ \cline{2-6}
\multicolumn{1}{r|}{} & m (items) & 1700 & 4000 & 100 & 27000
\\ \hline
\multicolumn{1}{|c|}{\multirow{4}{*}{\begin{tabular}[c]{@{}c@{}}Training Time \\ (sec/iteration)\end{tabular}}} & CMF & 4.80 & 44.14 & 126.98 & 967.840
\\ \cline{2-6}
\multicolumn{1}{|c|}{} & CPMF & 3.24 & 28.78 & 74.90 & 578.903
\\ \cline{2-6}
\multicolumn{1}{|c|}{} & GMF & 1.170 & 3.946 & - & -
\\ \cline{2-6}
\multicolumn{1}{|c|}{} & BGMF & 1.120 & 3.391 & 9.825 & 84.521
\\ \hline
\multicolumn{1}{|c|}{\multirow{4}{*}{Test RMSE}} & CMF & 0.955 & 0.921 & 2.163 & 0.862
\\ \cline{2-6}
\multicolumn{1}{|c|}{} & CPMF & 0.960 & 0.928 & 2.173 & 0.871
\\ \cline{2-6}
\multicolumn{1}{|c|}{} & GMF & 0.961 & 0.931 & - & -
\\ \cline{2-6}
\multicolumn{1}{|c|}{} & BGMF & 0.958 & 0.926 & 2.172 & 0.868\\ \hline
\end{tabular}

}

\caption[t3]{Comparison of CPU MF (CMF), Multi-threaded CPU MF (CPMF), Atomic GPU MF (GPMF), BGMF against standard data sets. Atomic variant of GPU MF could not be run on 4M Jester and MovieLense 20M data sets, due to short fall of GPU memory under constrained hardware (max memory allowed on each processing unit) taken for experimentation.}
\label{tab:standard}
\end{table}
It is observed from the values in Table~\ref{tab:standard} that the RMSE convergence of all variants are nearly equal (within the acceptable delta). The data sets used are limited to 4M size so it is possible to load the entire matrix into GPU and compare all 4 variants. However the GMF could not be run on large data sets, hence the related cells in Table ~\ref{tab:standard} are left blank (-). Additional hardware (with additional memory) can be make it possible to achieve GPU based synchronized factorization. However, BGMF could be run on Jester data set as the data needed for BGMF is limited to the specific blocks computed in parallel and not the entire data matrix. BGMF can be implemented for data sets of any size by limiting the size of the blocks, increasing the number of blocks in the block matrix and computing in parallel a subset of blocks. The underlying kernel implementation of all 4 variants are the same and the focus of these experiments is to evaluate performance of BGMF against other approaches. Tuning the parameters of underlying SGD kernel implementation to match the state-of-the-art algorithms is not an objective in this work. \\

\subsection{Block level (approximate) factorization}\label{sssec:block}

The set of experiments under this section are conducted  on a pseudo data set (generated with random numbers with values ranging between 1 to 30 ) of $1024\times1024$ matrix. The $\alpha=0.0001$, $\beta=0.01$ are the learning and regularization parameters with $\delta=0.01$ as the minimum difference of error between steps considered for early termination. The $k$ (number of latent features approximated) taken for the experiments is 10. The global optimum convergence RMSE value of the entire matrix is observed as 8.568 for the data set taken. The data matrix is split into blocks and processed on GPU with BGMF.

Rows 1-4 of Table~\ref{tab:variable} demonstrate the deviation in RMSE convergence with different number of iterations of factorization at a block level. It is observed that the convergence moves away from the global optima with each additional iteration of factorization at block level, even though the overall number of iterations (block level iterations $\times$ steps) increase. Rows 5-8 of Table~\ref{tab:variable} demonstrate the RMSE convergence with variable $G$ (number of iterations of factorization at a block level) which is incremented with every iteration. It is observed that the convergence moves closer to the optimum as we gradually increase the value of G during that step. The results for rows 5-7 shown in the table are obtained with $G$ value $<8$ to ensure that the individual blocks are not fully converged. Choice of large $G$ value leads to full convergence of the blocks with results similar to row 8. Row 8 of Table~\ref{tab:variable} demonstrates RMSE convergence with each block factorized till it converges. It is observed that the convergence of the entire matrix moves away from the optimal value as the final block's factors significantly impact the overall outcome. Row 9 of Table~\ref{tab:variable} demonstrates the RMSE convergence with variable number of iterations, at a block level, decremented from high value of 8 to low value of 1. It is observed that the convergence matches the global optimum when the high value is $<=8$ for the given data set. As the high value is $>8$, each block reaches full convergence causing challenges in transfer of latent features between blocks similar to row 9. The gradual decrease can be achieved by proportionally adjusting $G$ w.r.t to the ratio of RMSE improvement achieved in previous step.

\begin{table*}[h]
\caption[t4]{Variable block level convergence}
\label{tab:variable}
\begin{tabular}{|c|c|c|c|c|c|c|c|c|c|c|}
 \hline
 Steps --$>$ & 1 & 2 & 3 & 4 & 5 & 6 & 7 & 8 & 9 & 10 \\ \hline
 1 iteration/block & 9.217 & 8.779 & 8.631 & 8.587 & 8.575 & 8.571 & 8.569 & 8.569 & 8.569 & 8.569 \\
 \hline
 2 iterations/block & 8.766 & 8.586 & 8.572 & 8.570 & 8.570 & 8.570 & 8.570 & 8.570 & 8.570 & 8.570\\ \hline
 4 iterations/block & 8.598 & 8.578 & 8.575 & 8.576 & 8.576 & 8.576 & 8.576 & 8.576 & 8.576 & 8.576 \\ \hline
 8 iterations/block & 8.723 & 8.594 & 8.594 & 8.594 & 8.594 & 8.594 & 8.594 & 8.594 & 8.594 & 8.594 \\ \hline
 1,2,3..G iterations/block & 9.217 & 8.779 & 8.586 & 8.572 & 8.572 & 8.572 & 8.572 & 8.572 & 8.572 & 8.572 \\ \hline
 1,1,2..G/2 iterations/block & 9.217 & 8.779 & 8.631 & 8.587 & 8.571 & 8.570 & 8.571 & 8.571 & 8.571 & 8.571 \\ \hline
 1,1,1,1,2..G/3 iterations/block & 9.217 & 8.779 & 8.631 & 8.587 & 8.575 & 8.571 & 8.569 & 8.569 & 8.569 & 8.569 \\ \hline
 8...3,2,1,1,1 iterations/block & 8.723 & 8.589 & 8.585 & 8.581 & 8.577 & 8.574 & 8.572 & 8.571 & 8.570 & 8.569 \\ \hline
 Converge each block & 8.845 & 8.637 & 8.637 & 8.637 & 8.637 & 8.637 & 8.637 & 8.637 & 8.637 & 8.637 \\ \hline
 \end{tabular}
\end{table*}

\section{Conclusions}

Proposed approach of BGMF provides computational advantage, with respect to time and memory, without compromising on RMSE error convergence. By identifying optimal combination of GPU\_STEPS, CPU\_STEPS for BGMF, suitable for the specific data set, it is possible to achieve quicker convergence. Further performance gain is possible with variable block level iterations with approximate (but nearly optimal) convergence as detailed in Section \ref{sssec:block} (Block level factorization). By increasing the number of blocks, and by running in parallel exact number of blocks as the number of available cores, it is possible to factorize data sets of any scale. The technique can be adapted to run block level factorization on multiple GPUs as well as on distributed systems.\\

BGMF fundamentally does not change the underlying MF kernal itself, but merely makes it possible to achieve MF with block based approach. The technique also does not put limitation on the size of the data set. The size blocks are only limited by the available memory on the computation unit, while there is no limitation on the total number of blocks. Although we focused mostly on BGMF with equivalent block sizes, further experimentation is possible with variable block sizes and variable CPU, GPU iteration combinations at different phases of convergence. BGMF can further be enhanced to make it a divide and conquer approach with block level convergence, by identification of error limits at block level and by combining the results from multiple blocks. Further research is also possible in the direction of re-ordering the data elements so as to improve the efficiency of BGMF. \\

The technique could be combined with NNMF~\cite{Lee:2000:ANM:3008751.3008829}, Maximum Margin Matrix Factorization (MMMF)~\cite{Srebro:2004:MMF:2976040.2976207}, Probabilistic Matrix Factorization (PMF)~\cite{Salakhutdinov:2007:PMF:2981562.2981720} and any other variants of MF just by changing the kernel implementation within GPU. Although the experiments discussed largely focused on factorization of dense matrices, the approach has been extended to factorization of sparse matrices, as in the case of standard data sets, just by modifying the underlying GPU algorithm to work on sparse matrices and similar experimental results were noted. The approach can also be extended to approximate ALS approaches, that work on partial data, as implemented in~\cite{Tan2018MatrixFO} instead of SGD. The approach can be extended to transfer learning~\cite{DBLP:conf/premi/VPP17},~\cite{5288526} and ~\cite{Li20092052} where each domain contains limited data and combining data from both domains is applied to improve the overall recommendations. In this case, each block can represents the data specific to a domain and converged together using block MF. The approach can be applied to various domains (that leverage MF) ranging from recommender systems in collaborative filtering \cite{5197422} to text mining~\cite{10.1007/978-3-642-31178-9_15}. \\

\bibliographystyle{ACM-Reference-Format}
\bibliography{bgmf}


\begin{thebibliography}{23}


\ifx \showCODEN    \undefined \def \showCODEN     #1{\unskip}     \fi
\ifx \showDOI      \undefined \def \showDOI       #1{#1}\fi
\ifx \showISBNx    \undefined \def \showISBNx     #1{\unskip}     \fi
\ifx \showISBNxiii \undefined \def \showISBNxiii  #1{\unskip}     \fi
\ifx \showISSN     \undefined \def \showISSN      #1{\unskip}     \fi
\ifx \showLCCN     \undefined \def \showLCCN      #1{\unskip}     \fi
\ifx \shownote     \undefined \def \shownote      #1{#1}          \fi
\ifx \showarticletitle \undefined \def \showarticletitle #1{#1}   \fi
\ifx \showURL      \undefined \def \showURL       {\relax}        \fi
\providecommand\bibfield[2]{#2}
\providecommand\bibinfo[2]{#2}
\providecommand\natexlab[1]{#1}
\providecommand\showeprint[2][]{arXiv:#2}

\bibitem[Chen et~al\mbox{.}(2016)]%
        {7727641}
\bibfield{author}{\bibinfo{person}{W. Chen}, \bibinfo{person}{Y. Li},
  \bibinfo{person}{B. Pan}, {and} \bibinfo{person}{C. Xu}.}
  \bibinfo{year}{2016}\natexlab{}.
\newblock \showarticletitle{Block kernel Non-negative Matrix Factorization and
  its application to Face Recognition}. In \bibinfo{booktitle}{\emph{2016
  International Joint Conference on Neural Networks (IJCNN)}}.
  \bibinfo{pages}{3446--3452}.
\newblock
\showISSN{2161-4407}
\urldef\tempurl%
\url{https://doi.org/10.1109/IJCNN.2016.7727641}
\showDOI{\tempurl}


\bibitem[Chin et~al\mbox{.}(2015)]%
        {DBLP:conf/pakdd/ChinZJL15}
\bibfield{author}{\bibinfo{person}{Wei{-}Sheng Chin}, \bibinfo{person}{Yong
  Zhuang}, \bibinfo{person}{Yu{-}Chin Juan}, {and} \bibinfo{person}{Chih{-}Jen
  Lin}.} \bibinfo{year}{2015}\natexlab{}.
\newblock \showarticletitle{A Learning-Rate Schedule for Stochastic Gradient
  Methods to Matrix Factorization}. In \bibinfo{booktitle}{\emph{Advances in
  Knowledge Discovery and Data Mining - 19th Pacific-Asia Conference, {PAKDD}
  2015, Ho Chi Minh City, Vietnam, May 19-22, 2015, Proceedings, Part {I}}}.
  \bibinfo{pages}{442--455}.
\newblock
\urldef\tempurl%
\url{https://doi.org/10.1007/978-3-319-18038-0\_35}
\showDOI{\tempurl}


\bibitem[Du et~al\mbox{.}(2017)]%
        {Du2017}
\bibfield{author}{\bibinfo{person}{Rundong Du}, \bibinfo{person}{Da Kuang},
  \bibinfo{person}{Barry Drake}, {and} \bibinfo{person}{Haesun Park}.}
  \bibinfo{year}{2017}\natexlab{}.
\newblock \showarticletitle{DC-NMF: nonnegative matrix factorization based on
  divide-and-conquer for fast clustering and topic modeling}.
\newblock \bibinfo{journal}{\emph{Journal of Global Optimization}}
  \bibinfo{volume}{68}, \bibinfo{number}{4} (\bibinfo{date}{01 Aug}
  \bibinfo{year}{2017}), \bibinfo{pages}{777--798}.
\newblock
\showISSN{1573-2916}
\urldef\tempurl%
\url{https://doi.org/10.1007/s10898-017-0515-z}
\showDOI{\tempurl}


\bibitem[Gemulla et~al\mbox{.}(2011)]%
        {DBLP:conf/kdd/GemullaNHS11}
\bibfield{author}{\bibinfo{person}{Rainer Gemulla}, \bibinfo{person}{Erik
  Nijkamp}, \bibinfo{person}{Peter~J. Haas}, {and} \bibinfo{person}{Yannis
  Sismanis}.} \bibinfo{year}{2011}\natexlab{}.
\newblock \showarticletitle{Large-scale matrix factorization with distributed
  stochastic gradient descent}. In \bibinfo{booktitle}{\emph{Proceedings of the
  17th {ACM} {SIGKDD} International Conference on Knowledge Discovery and Data
  Mining, San Diego, CA, USA, August 21-24, 2011}}. \bibinfo{pages}{69--77}.
\newblock
\urldef\tempurl%
\url{https://doi.org/10.1145/2020408.2020426}
\showDOI{\tempurl}


\bibitem[Koitka and Friedrich(2016)]%
        {koitka2016nmfgpu4r}
\bibfield{author}{\bibinfo{person}{Sven Koitka} {and}
  \bibinfo{person}{Christoph~M Friedrich}.} \bibinfo{year}{2016}\natexlab{}.
\newblock \showarticletitle{nmfgpu4R: GPU-accelerated computation of the
  Non-Negative Matrix Factorization (NMF) using CUDA capable hardware}.
\newblock \bibinfo{journal}{\emph{The R Journal}} \bibinfo{volume}{8},
  \bibinfo{number}{2} (\bibinfo{year}{2016}), \bibinfo{pages}{382--392}.
\newblock


\bibitem[Koren et~al\mbox{.}(2009)]%
        {5197422}
\bibfield{author}{\bibinfo{person}{Y. Koren}, \bibinfo{person}{R. Bell}, {and}
  \bibinfo{person}{C. Volinsky}.} \bibinfo{year}{2009}\natexlab{}.
\newblock \showarticletitle{Matrix Factorization Techniques for Recommender
  Systems}.
\newblock \bibinfo{journal}{\emph{Computer}} \bibinfo{volume}{42},
  \bibinfo{number}{8} (\bibinfo{date}{Aug} \bibinfo{year}{2009}),
  \bibinfo{pages}{30--37}.
\newblock
\showISSN{0018-9162}
\urldef\tempurl%
\url{https://doi.org/10.1109/MC.2009.263}
\showDOI{\tempurl}


\bibitem[Kysenko et~al\mbox{.}(2012)]%
        {10.1007/978-3-642-31178-9_15}
\bibfield{author}{\bibinfo{person}{Volodymyr Kysenko}, \bibinfo{person}{Karl
  Rupp}, \bibinfo{person}{Oleksandr Marchenko}, \bibinfo{person}{Siegfried
  Selberherr}, {and} \bibinfo{person}{Anatoly Anisimov}.}
  \bibinfo{year}{2012}\natexlab{}.
\newblock \showarticletitle{GPU-Accelerated Non-negative Matrix Factorization
  for Text Mining}. In \bibinfo{booktitle}{\emph{Natural Language Processing
  and Information Systems}}, \bibfield{editor}{\bibinfo{person}{Gosse Bouma},
  \bibinfo{person}{Ashwin Ittoo}, \bibinfo{person}{Elisabeth Mtais}, {and}
  \bibinfo{person}{Hans Wortmann}} (Eds.). \bibinfo{publisher}{Springer Berlin
  Heidelberg}, \bibinfo{address}{Berlin, Heidelberg},
  \bibinfo{pages}{158--163}.
\newblock
\showISBNx{978-3-642-31178-9}


\bibitem[Lee and Seung(2000)]%
        {Lee:2000:ANM:3008751.3008829}
\bibfield{author}{\bibinfo{person}{Daniel~D. Lee} {and}
  \bibinfo{person}{H.~Sebastian Seung}.} \bibinfo{year}{2000}\natexlab{}.
\newblock \showarticletitle{Algorithms for Non-negative Matrix Factorization}.
  In \bibinfo{booktitle}{\emph{Proceedings of the 13th International Conference
  on Neural Information Processing Systems}} (Denver, CO)
  \emph{(\bibinfo{series}{NIPS'00})}. \bibinfo{publisher}{MIT Press},
  \bibinfo{address}{Cambridge, MA, USA}, \bibinfo{pages}{535--541}.
\newblock
\urldef\tempurl%
\url{http://dl.acm.org/citation.cfm?id=3008751.3008829}
\showURL{%
\tempurl}


\bibitem[Li et~al\mbox{.}(2013)]%
        {DBLP:conf/edbt/LiTS13}
\bibfield{author}{\bibinfo{person}{Boduo Li}, \bibinfo{person}{Sandeep Tata},
  {and} \bibinfo{person}{Yannis Sismanis}.} \bibinfo{year}{2013}\natexlab{}.
\newblock \showarticletitle{Sparkler: supporting large-scale matrix
  factorization}. In \bibinfo{booktitle}{\emph{Joint 2013 {EDBT/ICDT}
  Conferences, {EDBT} '13 Proceedings, Genoa, Italy, March 18-22, 2013}}.
  \bibinfo{pages}{625--636}.
\newblock
\urldef\tempurl%
\url{https://doi.org/10.1145/2452376.2452449}
\showDOI{\tempurl}


\bibitem[Li et~al\mbox{.}(2009)]%
        {Li20092052}
\bibfield{author}{\bibinfo{person}{B. Li}, \bibinfo{person}{Q. Yang}, {and}
  \bibinfo{person}{X. Xue}.} \bibinfo{year}{2009}\natexlab{}.
\newblock \showarticletitle{Can movies and books collaborate? Cross-domain
  collaborative filtering for sparsity reduction}.
\newblock \bibinfo{journal}{\emph{IJCAI International Joint Conference on
  Artificial Intelligence}} (\bibinfo{year}{2009}),
  \bibinfo{pages}{2052--2057}.
\newblock
\showISBNx{9781577354260}
\showISSN{10450823}
\newblock
\shownote{cited By 186}.


\bibitem[Mackey et~al\mbox{.}(2011)]%
        {NIPS20114486}
\bibfield{author}{\bibinfo{person}{Lester~W. Mackey},
  \bibinfo{person}{Michael~I. Jordan}, {and} \bibinfo{person}{Ameet
  Talwalkar}.} \bibinfo{year}{2011}\natexlab{}.
\newblock \showarticletitle{Divide-and-Conquer Matrix Factorization}.
\newblock In \bibinfo{booktitle}{\emph{Advances in Neural Information
  Processing Systems 24}}, \bibfield{editor}{\bibinfo{person}{J.~Shawe-Taylor},
  \bibinfo{person}{R.~S. Zemel}, \bibinfo{person}{P.~L. Bartlett},
  \bibinfo{person}{F.~Pereira}, {and} \bibinfo{person}{K.~Q. Weinberger}}
  (Eds.). \bibinfo{publisher}{Curran Associates, Inc.},
  \bibinfo{pages}{1134--1142}.
\newblock
\urldef\tempurl%
\url{http://papers.nips.cc/paper/4486-divide-and-conquer-matrix-factorization.pdf}
\showURL{%
\tempurl}


\bibitem[Oh et~al\mbox{.}(2015)]%
        {DBLP:conf/kdd/OhHYJ15}
\bibfield{author}{\bibinfo{person}{Jinoh Oh}, \bibinfo{person}{Wook{-}Shin
  Han}, \bibinfo{person}{Hwanjo Yu}, {and} \bibinfo{person}{Xiaoqian Jiang}.}
  \bibinfo{year}{2015}\natexlab{}.
\newblock \showarticletitle{Fast and Robust Parallel {SGD} Matrix
  Factorization}. In \bibinfo{booktitle}{\emph{Proceedings of the 21th {ACM}
  {SIGKDD} International Conference on Knowledge Discovery and Data Mining,
  Sydney, NSW, Australia, August 10-13, 2015}}. \bibinfo{pages}{865--874}.
\newblock
\urldef\tempurl%
\url{https://doi.org/10.1145/2783258.2783322}
\showDOI{\tempurl}


\bibitem[Pan and Yang(2010)]%
        {5288526}
\bibfield{author}{\bibinfo{person}{S.~J. Pan} {and} \bibinfo{person}{Q. Yang}.}
  \bibinfo{year}{2010}\natexlab{}.
\newblock \showarticletitle{A Survey on Transfer Learning}.
\newblock \bibinfo{journal}{\emph{IEEE Transactions on Knowledge and Data
  Engineering}} \bibinfo{volume}{22}, \bibinfo{number}{10} (\bibinfo{date}{Oct}
  \bibinfo{year}{2010}), \bibinfo{pages}{1345--1359}.
\newblock
\showISSN{1041-4347}
\urldef\tempurl%
\url{https://doi.org/10.1109/TKDE.2009.191}
\showDOI{\tempurl}


\bibitem[Recht et~al\mbox{.}(2011)]%
        {DBLP:conf/nips/RechtRWN11}
\bibfield{author}{\bibinfo{person}{Benjamin Recht},
  \bibinfo{person}{Christopher R{\'{e}}}, \bibinfo{person}{Stephen~J. Wright},
  {and} \bibinfo{person}{Feng Niu}.} \bibinfo{year}{2011}\natexlab{}.
\newblock \showarticletitle{Hogwild: {A} Lock-Free Approach to Parallelizing
  Stochastic Gradient Descent}. In \bibinfo{booktitle}{\emph{Advances in Neural
  Information Processing Systems 24: 25th Annual Conference on Neural
  Information Processing Systems 2011. Proceedings of a meeting held 12-14
  December 2011, Granada, Spain.}} \bibinfo{pages}{693--701}.
\newblock


\bibitem[Salakhutdinov and Mnih(2007)]%
        {Salakhutdinov:2007:PMF:2981562.2981720}
\bibfield{author}{\bibinfo{person}{Ruslan Salakhutdinov} {and}
  \bibinfo{person}{Andriy Mnih}.} \bibinfo{year}{2007}\natexlab{}.
\newblock \showarticletitle{Probabilistic Matrix Factorization}. In
  \bibinfo{booktitle}{\emph{Proceedings of the 20th International Conference on
  Neural Information Processing Systems}} (Vancouver, British Columbia, Canada)
  \emph{(\bibinfo{series}{NIPS'07})}. \bibinfo{publisher}{Curran Associates
  Inc.}, \bibinfo{address}{USA}, \bibinfo{pages}{1257--1264}.
\newblock
\showISBNx{978-1-60560-352-0}


\bibitem[Schelter et~al\mbox{.}(2014)]%
        {DBLP:journals/corr/SchelterSZ14}
\bibfield{author}{\bibinfo{person}{Sebastian Schelter}, \bibinfo{person}{Venu
  Satuluri}, {and} \bibinfo{person}{Reza Zadeh}.}
  \bibinfo{year}{2014}\natexlab{}.
\newblock \showarticletitle{Factorbird - a Parameter Server Approach to
  Distributed Matrix Factorization}.
\newblock \bibinfo{journal}{\emph{CoRR}}  \bibinfo{volume}{abs/1411.0602}
  (\bibinfo{year}{2014}).
\newblock
\showeprint[arxiv]{1411.0602}
\urldef\tempurl%
\url{http://arxiv.org/abs/1411.0602}
\showURL{%
\tempurl}


\bibitem[Srebro et~al\mbox{.}(2004)]%
        {Srebro:2004:MMF:2976040.2976207}
\bibfield{author}{\bibinfo{person}{Nathan Srebro}, \bibinfo{person}{Jason D.~M.
  Rennie}, {and} \bibinfo{person}{Tommi~S. Jaakkola}.}
  \bibinfo{year}{2004}\natexlab{}.
\newblock \showarticletitle{Maximum-margin Matrix Factorization}. In
  \bibinfo{booktitle}{\emph{Proceedings of the 17th International Conference on
  Neural Information Processing Systems}} (Vancouver, British Columbia, Canada)
  \emph{(\bibinfo{series}{NIPS'04})}. \bibinfo{publisher}{MIT Press},
  \bibinfo{address}{Cambridge, MA, USA}, \bibinfo{pages}{1329--1336}.
\newblock


\bibitem[Tan et~al\mbox{.}(2018)]%
        {Tan2018MatrixFO}
\bibfield{author}{\bibinfo{person}{Wei Tan}, \bibinfo{person}{Shiyu Chang},
  \bibinfo{person}{Liana~L. Fong}, \bibinfo{person}{Cheng Li},
  \bibinfo{person}{Zijun Wang}, {and} \bibinfo{person}{Liangliang Cao}.}
  \bibinfo{year}{2018}\natexlab{}.
\newblock \showarticletitle{Matrix Factorization on GPUs with Memory
  Optimization and Approximate Computing}. In \bibinfo{booktitle}{\emph{ICPP}}.
\newblock


\bibitem[V. et~al\mbox{.}(2017)]%
        {DBLP:conf/premi/VPP17}
\bibfield{author}{\bibinfo{person}{Sowmini~Devi V.}, \bibinfo{person}{Vineet
  Padmanabhan}, {and} \bibinfo{person}{Arun~K. Pujari}.}
  \bibinfo{year}{2017}\natexlab{}.
\newblock \showarticletitle{A Matrix Factorization and Clustering Based
  Approach for Transfer Learning}. In \bibinfo{booktitle}{\emph{Pattern
  Recognition and Machine Intelligence - 7th International Conference,PReMI
  2017, Kolkata, India, December 5-8, 2017, Proceedings}}.
  \bibinfo{pages}{77--83}.
\newblock
\urldef\tempurl%
\url{https://doi.org/10.1007/978-3-319-69900-4\_10}
\showDOI{\tempurl}


\bibitem[Yu et~al\mbox{.}(2012)]%
        {DBLP:conf/icdm/YuHSD12}
\bibfield{author}{\bibinfo{person}{Hsiang{-}Fu Yu}, \bibinfo{person}{Cho{-}Jui
  Hsieh}, \bibinfo{person}{Si Si}, {and} \bibinfo{person}{Inderjit~S.
  Dhillon}.} \bibinfo{year}{2012}\natexlab{}.
\newblock \showarticletitle{Scalable Coordinate Descent Approaches to Parallel
  Matrix Factorization for Recommender Systems}. In
  \bibinfo{booktitle}{\emph{12th {IEEE} International Conference on Data
  Mining, {ICDM} 2012, Brussels, Belgium, December 10-13, 2012}}.
  \bibinfo{pages}{765--774}.
\newblock
\urldef\tempurl%
\url{https://doi.org/10.1109/ICDM.2012.168}
\showDOI{\tempurl}


\bibitem[Yun et~al\mbox{.}(2014)]%
        {Yun:2014:NNS:2732967.2732973}
\bibfield{author}{\bibinfo{person}{Hyokun Yun}, \bibinfo{person}{Hsiang-Fu Yu},
  \bibinfo{person}{Cho-Jui Hsieh}, \bibinfo{person}{S.~V.~N. Vishwanathan},
  {and} \bibinfo{person}{Inderjit Dhillon}.} \bibinfo{year}{2014}\natexlab{}.
\newblock \showarticletitle{NOMAD: Non-locking, Stochastic Multi-machine
  Algorithm for Asynchronous and Decentralized Matrix Completion}.
\newblock \bibinfo{journal}{\emph{Proc. VLDB Endow.}} \bibinfo{volume}{7},
  \bibinfo{number}{11} (\bibinfo{date}{July} \bibinfo{year}{2014}),
  \bibinfo{pages}{975--986}.
\newblock
\showISSN{2150-8097}
\urldef\tempurl%
\url{https://doi.org/10.14778/2732967.2732973}
\showDOI{\tempurl}


\bibitem[Zhang et~al\mbox{.}(2013)]%
        {Zhang:2013:LMF:2488388.2488520}
\bibfield{author}{\bibinfo{person}{Yongfeng Zhang}, \bibinfo{person}{Min
  Zhang}, \bibinfo{person}{Yiqun Liu}, \bibinfo{person}{Shaoping Ma}, {and}
  \bibinfo{person}{Shi Feng}.} \bibinfo{year}{2013}\natexlab{}.
\newblock \showarticletitle{Localized Matrix Factorization for Recommendation
  Based on Matrix Block Diagonal Forms}. In
  \bibinfo{booktitle}{\emph{Proceedings of the 22Nd International Conference on
  World Wide Web}} (Rio de Janeiro, Brazil) \emph{(\bibinfo{series}{WWW '13})}.
  \bibinfo{publisher}{ACM}, \bibinfo{address}{New York, NY, USA},
  \bibinfo{pages}{1511--1520}.
\newblock
\showISBNx{978-1-4503-2035-1}
\urldef\tempurl%
\url{https://doi.org/10.1145/2488388.2488520}
\showDOI{\tempurl}


\bibitem[Zhuang et~al\mbox{.}(2013)]%
        {Zhuang:2013:FPS:2507157.2507164}
\bibfield{author}{\bibinfo{person}{Yong Zhuang}, \bibinfo{person}{Wei-Sheng
  Chin}, \bibinfo{person}{Yu-Chin Juan}, {and} \bibinfo{person}{Chih-Jen Lin}.}
  \bibinfo{year}{2013}\natexlab{}.
\newblock \showarticletitle{A Fast Parallel SGD for Matrix Factorization in
  Shared Memory Systems}. In \bibinfo{booktitle}{\emph{Proceedings of the 7th
  ACM Conference on Recommender Systems}} (Hong Kong, China)
  \emph{(\bibinfo{series}{RecSys '13})}. \bibinfo{publisher}{ACM},
  \bibinfo{address}{New York, NY, USA}, \bibinfo{pages}{249--256}.
\newblock
\showISBNx{978-1-4503-2409-0}
\urldef\tempurl%
\url{https://doi.org/10.1145/2507157.2507164}
\showDOI{\tempurl}


\end{thebibliography}

\appendix

\end{document}